\documentclass{article}

\usepackage[american]{babel}
\usepackage{avm}
\usepackage[sort]{natbib}
\usepackage{pstricks}
\usepackage{pst-tree}

\avmfont{\sc}
\avmvskip{0ex}

\psset{treemode=U,treesep=.1cm,levelsep=1cm}

\newcommand{\mathsc}[1]{\mbox{\sc #1}}

\begin{document}

\title{Applying Constraint Handling Rules to HPSG}

\author{Gerald Penn\\
        Eberhard-Karls-Universit\"at T\"ubingen and Bell Laboratories}
\date{June 30, 2000}

\maketitle

\begin{abstract}
Constraint Handling Rules (CHR) have provided a realistic solution to
an over-arching problem in many fields that deal with constraint logic
programming: how to combine recursive functions or relations with
constraints while avoiding non-termination problems. This paper
focuses on some other benefits that CHR, specifically their
implementation in SICStus Prolog, have provided to computational
linguists working on grammar design tools.  CHR rules are applied by
means of a subsumption check and this check is made only when
their variables are instantiated or bound.  The former functionality
is at best difficult to simulate using more primitive coroutining
statements such as SICStus when/2, and the latter simply did not exist
in any form before CHR.

For the sake of providing a case study in how these can be applied to
grammar development, we consider the Attribute Logic Engine (ALE), a
Prolog preprocessor for logic programming with typed feature
structures, and its extension to a complete grammar development system
for Head-driven Phrase Structure Grammar (HPSG), a popular
constraint-based linguistic theory that uses typed feature structures.
In this context, CHR can be used not only to extend the constraint
language of feature structure descriptions to include relations in a
declarative way, but also to provide support for constraints with
complex antecedents and constraints on the co-occurrence of feature
values that are necessary to interpret the type system of HPSG
properly.
\end{abstract}

Constraint Handling Rules \citep[CHR,\ ][]{fruehwirth-abdennadher97}
have provided a realistic solution to an over-arching problem in many
fields that deal with constraint logic programming: how to combine
recursive functions or relations with constraints while avoiding
non-termination problems.  Potential applications of constraint logic
programming within computational linguistics have certainly not been
immune to this problem, and for them, CHR is currently the only
existing tractable alternative to manually specifying a set of delay
statements to avoid non-termination in a space of constraints with
huge numbers of non-deterministic constraints and relational
attachments that are almost invariably recursive.

This paper focuses on some other benefits that CHR, specifically its
implementation in SICStus Prolog, has provided to computational
linguists working on grammar design tools.  CHR rules are applied
through a subsumption check and this check is made only when their
variables are instantiated or bound.  The former functionality is at
best difficult to simulate using more primitive coroutining statements
such as SICStus \texttt{when/2}, and the latter simply did not exist
in any form before CHR, to the best of this author's knowledge.

For the sake of providing a case study in how these can be applied to
grammar development, we will consider the Attribute Logic Engine (ALE)
\citep{carp-penn96}, a Prolog preprocessor for logic programming over
the logic of typed feature structures \citep{carpenter92}, and its
extension to a complete grammar development system for Head-driven
Phrase Structure Grammar (HPSG) \citep{pollard-sag87,pollard-sag94}, a
popular constraint-based linguistic theory that uses typed feature
structures.  A great many details of HPSG have changed since its
inception and still more are yet to be agreed upon, so the
presentation here is by necessity a tremendous simplification.
Nevertheless, it is hoped that this paper will stimulate a discussion
of what limit CHR permits us to extend the state of the art in grammar
development to, and what future developments on the part of
implementors of logic programming languages such as Prolog would be
most beneficial in order to push that limit even further.

The next three sections give a very brief introduction to the view of
human languages that HPSG takes, how typed feature structures fit into
that view, and what the basic constraint-solving problem is that
grammar developers are faced with.  Section~\ref{decomp:sec} then
shows how to compile these general constraints into more atomic ones
and where CHR fills in the gap in providing an elegant implementation
of this in Prolog.  Section~\ref{subcover:sec} illustrates another
kind of constraint that must be dealt with internally by HPSG grammar
development systems, and how CHR provides the unique ability to do
that efficiently.

Finally, Section~\ref{desiderata:sec} points to a few issues that
still remain to be resolved relative to what Prolog with its
CLP extensions is currently able to provide, namely a more
efficient implementation of attributed variables and a means for
universal quantification of constraints.

\section{Head-driven Phrase Structure Grammar}

HPSG is a theory that seeks to predict the grammaticality, or
syntactic well-formedness, of utterances in any human language.  It is
sometimes called a \emph{realist} theory of language, as opposed to a
\emph{mentalist} theory of language, because grammatical utterances
from human languages are taken to be actual entities that exist in the
world.  As such, these entities constitute a universe of objects for a
model that we seek to describe using HPSG.  The principal role of a
grammar development system in this context is to provide the means to
state theories of human language formally, and then to ask whether the
formal representation of a candidate utterance with a particular
phonological realization, i.e., one comprised of a particular string
of words, has a non-empty denotation in this model.

By convention, the theory itself is stated using the formal language
of the logic of typed feature structures.  Typed feature structures
are a semantically typed kind of record, which is a very convenient
representation for linguistic purposes because the combination of
typing with named attributes allows for a very terse description
language that can easily make reference to a sparse amount of
information in what are usually extremely large structures/records.
There are actually several variations of that language in circulation,
e.g., \citealp{smolka88,king89,carpenter92}, which are more or less
well-suited to the purposes of HPSG, or, perhaps more historically
accurate, to the rules of which linguists working with HPSG more or
less elected to adhere.  The language that the seminal work on HPSG of
\citet{pollard-sag94} actually uses was finally formalized in its own
right by \citet{rsp99,richter00}.  The bad news is that satisfiability
in this language is not even compact.  The good news is that no one
particularly seems to care about the bad news.  This is due to the
fact that: (1) most of the work performed by linguists in HPSG is not
actually formalized in this language, thus creating a gap between
theory and practice that must be bridged in the course of
formalization in a grammar development system anyway; and (2) most
grammar development systems have, for years, elected to follow a much
more conservative route inspired by work in constraint logic
programming, in which a core constraint language for describing typed
feature structures has been extended with a relational definite clause
language \citep{hoehfeld-smolka88} for conveniently stating the
``constraints,'' or as they are called in linguistics,
\emph{principles}, that define grammaticality in human language.  The
core constraint language that will be assumed here is essentially the
one defined by \citet{carpenter92}, with an eye towards the view of
feature structures in HPSG taken by \citet{king99}.

\section{Typed Feature Structures and their Descriptions}

Each typed feature structure is a tuple $\langle
Q,\theta,\delta\rangle$, where $Q$ is a set of nodes (corresponding to
substructures of feature structures), $\theta$ is a total typing
function that maps nodes to a fixed finite meet semi-lattice of types,
\textit{Type}, and $\delta$ is a partial feature value function that
maps pairs consisting of a feature, drawn from a finite set,
\textit{Feat}, and a node in $Q$ to nodes in $Q$.  Typed feature
structures are conventionally depicted as records, such as:
\begin{center}
\begin{avm}
\[ \rm throwing \\
thrower\; \[ \rm index \\
             person\; \rm third\\
             number\; \@1 \rm singular\\
             gender\; \rm masculine
          \] \\
thrown\; \[ \rm index\\
            person\; \rm third\\
            number\; \@1\\
            gender\; \rm neuter
         \]
\]
\end{avm}
\end{center}
In this example, there is a total of eight nodes, one of type
\emph{throwing}, two of type \emph{index}, two of type \emph{third},
and one each of type \emph{singular}, \emph{masculine} and
\emph{neuter}.  Note that the numeric tag (called a
\emph{re-entrancy}) indicates that the two number features have
extensionally the same value, which is of type \emph{singular},
whereas the two person features have different feature structure
values, i.e., different nodes, that happen to be of the same type,
\emph{third}.  Typed feature logics are normally intensionally typed,
so this identity of type does not imply an extensional identity of
feature values.  We can refer to substructures of feature structures
along paths, provided that the values of each feature on the path
exist.  The \textsc{thrower}:\textsc{gender} value of this example is
of type \emph{masculine}.

The core constraint language itself is a description logic for talking
about feature structures relative to the finite meet semi-lattice of
types, \textit{Type}, a fixed finite set of features, \textit{Feat},
and a countable set of variables, \textit{Var}:
\begin{quote}
The description language, $\Phi$, is the least set of descriptions
that contains:
\begin{itemize}
\item $v, v\in \mathit{Var},$
\item $\tau, \tau\in \mathit{Type},$
\item $\mathsc{f}:\phi, \mathsc{f}\in \mathit{Feat},\phi\in\Phi,$
\item $\phi_1 \wedge \phi_2, \phi_1,\phi_2\in \Phi,$ and
\item $\phi_1 \vee \phi_2, \phi_1,\phi_2\in \Phi.$
\end{itemize}
\end{quote}
A nice property of this language is that every non-disjunctive
description with a non-empty denotation has a unique most general
feature structure in its denotation.  This is called its \emph{most
general satisfier}.  Of course, this is not the same kind of
denotation as the one that connects our theory to language itself ---
there are two levels of abstraction at work.  Descriptions denote sets
of feature structures, and feature structures denote sets of
utterances in the world.

A given set of types and features is also typically augmented with a
set of \emph{appropriateness conditions} that mediate their
occurrence, following \citet{king89} and \citet{carpenter92}.
Appropriateness conditions stipulate, for every type, a finite set of
features that can and must have values in feature structures of that
type.  This effectively forces feature structures to be
finite-branching terms with named attributes, although the
introduction of new features combined with subtyping means that the
arities of these terms are not completely fixed.  Appropriateness
conditions also usually restrict the value that appropriate features
can take by specifying a type that must be a supertype of the value's
type.

In the formulation of \citet{carpenter92}, appropriateness must also
guarantee that there is a unique most general type for which a given
feature is appropriate.  This is called \emph{unique feature
introduction}, and for feature \textsc{f}, its introducing type is
called $Intro(\mathsc{f})$.  Until recently, it was generally believed
that Prolog systems that work with typed feature structures must
meta-interpret unification and SLD resolution over them.  As shown by
\citet{penn00}, when unique feature introduction is assumed, every
statically typable set of appropriateness conditions admits a Prolog
term encoding of its typed feature structures, thus reducing feature
structure unification to Prolog unification in the underlying Warren
abstract machine.  When it is not assumed, furthermore, unique feature
introduction can be added automatically in polynomial time while
preserving meet-semi-latticehood. This result opens the door to using
pieces of Prolog implementations' enhanced functionality, including
CHR, that would have been prohibitively less elegant if even only
unification needed to be meta-interpreted.

\section{The Language of Principles}

Principles can be divided into two classes: those that apply to all
languages, and those that only apply to specific languages.  The
lexicon itself can be regarded as a very disjunctive principle ---
language-specific, of course --- associating each word to a specific
syntactic and semantic representation.  This paper will not say much
more about the lexicon.

In HPSG, and, in fact, in nearly every constraint-based theory of
grammar, all principles, regardless of class, take the form of a
statement, ``for all terms/feature structures, statement $\pi$
holds.''  In contrast to most work in constraint logic programming,
every constraint is thus implicitly universally quantified over typed
feature structures, and thus not explicitly posted to a store.  In
practice, nearly every principle's $\pi$ takes the form of an
implication, and for most complex principles, $\pi$ makes reference to
instances of relations.  The approach taken here is that all $\pi$ are
of the form:
\[
\alpha \Longrightarrow (\gamma \wedge \rho)
\]
where $\alpha$ and $\gamma$ are descriptions from the core constraint
language and $\rho$ is drawn from a definite clause language of
relations, whose arguments are also descriptions from the core
constraint language.  Typically, these arguments include variables
that are also used in $\gamma$, so that variables take scope over the
entire consequent.

This form is somewhat of an idealization itself.  Linguists often
formulate principles in which variables are shared between $\alpha$
and the consequent, with the intention that $\alpha \Longrightarrow
(\gamma \wedge \rho)$ should actually be $\alpha \Longrightarrow
(\alpha \wedge \gamma \wedge \rho)$.  Sometimes, they also use
relations in $\alpha$ rather than just descriptions from the core
language, for the interpretation of which negation by failure or some
other extra convention must be used.  These will be ignored in the
rest of this paper, although the method described below for dealing
with this form of principle can be extended in several ways to attempt
to handle these more complicated antecedents as well.

The presumed form of these principles is not new.  The ALE system
permits the same form, but where $\alpha$ can only be a type
description, $\tau$.  The ConTroll system \citep{goetz-meurers97}
permits essentially the same form, but interprets the implications
classically, which creates a very large search space with severe
non-termination problems because of the presence of potentially
recursively defined relations in $\rho$.  ConTroll attempted to remedy
this problem by providing a language of delays, similar to coroutining
predicates in Prolog but with the extra implicit universal
quantification built in.  Repeated attempts at large-scale development
with this method proved unpromising, as finding the right set of
delays actually proved to be more work than simply using a more
traditional deductive strategy such as SLD resolution or definite
clause grammars.  Such delays are also non-modular relative to changes
in other parts of the grammar, which is of particular concern since
grammar development is largely characterized by a large number of
small incremental changes.

By contrast, the approach taken here is to allow for arbitrary
$\alpha$ but to interpret the implications using subsumption by
$\alpha$, i.e., for every feature structure (the implicit universal
quantification is still there), either the consequent holds, or the
feature structure is not subsumed by the most general satisfier of
$\alpha$.  This decision draws its inspiration from three sources.
First, ALE used the same interpretation, although somewhat more
controversially given that $\alpha$ could only be a type.  Second, it
is widely believed in the linguistics community
\citep{hinrichs-nakazawa96} that \emph{lexical rules}, closure
conditions on the well-formedness of a lexicon, are intended to apply
by subsumption.  They also take the form of implications, as ``if
$\lambda_1$ is part of the lexicon, then so is $\lambda_2$,'' in which
$\lambda_1$ and $\lambda_2$ are specified using descriptions with or
without some extra level of default reasoning.  One can, in fact, also
find many instances of principles in theoretical linguistics
literature in which a subsumption-based interpretation is the intended
one.  The third source is the approach to CHR in its SICStus Prolog
implementation, in which the head of a rule is matched using a
subsumption check.  The fact that linguistic theory, conventional
grammar development practice and the CHR package agree on this point
is very significant, and is exploited by our implementation.

The subsumption-based approach is sound with respect to the classical
interpretation of implication for those cases of principles when the
classical interpretation really is the correct one.  For completeness,
other resolution methods must be used.  These are available from true
constraint logic programming, which as of yet is surprisingly absent
from linguistic research on constraint-based grammar.  CHR can provide
those additional resolution methods as well, of course.

\section{Principle Decomposition}
\label{decomp:sec}

Under these assumptions, every linguistic principle can be decomposed
into a series of more atomic constraints that can easily be
implemented.  It should first be noted that the existence of a CHR
package for Prolog terms easily admits a CHR package for typed feature
structures, where feature descriptions in the arguments of constraints
are translated to the term encodings of their most general satisfiers.
The result can then be compiled by the underlying Prolog CHR library's
compiler.  This immediately provides grammar developers with an
alternative to providing explicit delay statements that has proven to
be far more robust and tractable for large-scale grammar development.
The rest of this section concerns the compilation of the universally
quantified principles themselves.

\subsection{Finding a Trigger}

Let $\mathit{trigger}(\alpha)$ be defined such that:
\begin{itemize}
\item $\mathit{trigger}(v) = \bot$,
\item $\mathit{trigger}(\tau) = \tau$,
\item $\mathit{trigger}(\mathsc{f}:\phi) =
  \mathit{Intro}(\mathsc{f})$,
\item $\mathit{trigger}(\phi_1\wedge \phi_2) =
\mathit{trigger}(\phi_1) \sqcup \mathit{trigger}(\phi_2)$, and
\item $\mathit{trigger}(\phi_1\vee \phi_2) =
\mathit{trigger}(\phi_1) \sqcap \mathit{trigger}(\phi_2)$,
\end{itemize}
where $\bot$ is the most general type in \textit{Type}, and
$\mathord{\sqcup}$ and $\mathord{\sqcap}$ are join (unification)
and meet (generalization) in \textit{Type}.  $\bot$ and meet exist
given the assumption that \textit{Type} is a meet semi-lattice.  From
that same assumption, join exists between any pair of consistent
types.

If we had a predicate, \texttt{fswhen(V=Desc,Goal)}, available such that
we could delay \texttt{Goal} until \texttt{V} was subsumed by the most
general satisfier of \texttt{Desc}, then all principles:
\[
\alpha \Longrightarrow (\gamma \wedge \rho)
\]
could be converted into:
\[
\mathit{trigger}(\alpha) \Longrightarrow v \wedge
  \mathtt{fswhen}((v=\alpha), ((v=\gamma) \wedge \rho))
\]
This form (with the exception of \texttt{fswhen/2}) is the form already
implemented by ALE, including its implicit universal quantification.
This typically involves compiling unifications of feature structures
into unifications of their term encodings plus a type check to see
whether one or more principle consequents need to be applied.  The
conditions under which that check must be made can, of course, be
statically optimized to a bare minimum, but in the worst case, it
significantly compromises the speed of what would otherwise be a
simple Prolog-level unification.  This point is discussed further in
Section~\ref{desiderata:sec}.

As a running example, we can consider a principle from the HPSG
grammar for German under development at the University of T\"ubingen:
\begin{verbatim}
synsem:loc:cat:(head:verb, marking:fin) 
  ==> synsem:loc:cat:head:vform:bse.
\end{verbatim}
This is called the Finiteness Marking Principle.  The description to
the left of the implication selects feature structures whose
substructure on the path \textsc{synsem}:\textsc{loc}:\textsc{cat}
satisfy two requirements: that their \textsc{head} value is of the
type \emph{verb} and that their \textsc{marking} value is of the type
\emph{fin}.  The principle says that every feature structure that
satisfies that description must also have a \textsc{synsem}:
\textsc{loc}:\textsc{cat}:\textsc{head}:\textsc{vform} value of type
\emph{bse}.  The problem is that feature structures may not have
substructures that are specific enough to determine whether this
constraint holds or not.  This can happen when a type subsumes several
more specific subtypes of which only one meets the requirements set
forth by the constraint, for example.  So we must wait until it is
known whether the antecedent is true or false before applying the
consequent.  If we reach a deadlock, where several constraints are
suspended on their antecedents, then we must use another resolution
method to begin testing more specific extensions of the feature
structure in turn.

To find the trigger in this example, we can observe that the
antecedent is a feature value description (\textsc{f}:$\phi$), so the
trigger is \emph{Intro}(\textsc{synsem}), the unique introducer of the
\textsc{synsem} feature, which happens to be the type \emph{sign}.  We
can then transform this constraint to:
\begin{verbatim}
sign cons X 
     goal fswhen((X=synsem:loc:cat:(head:verb, marking:fin)),
                 (X=synsem:loc:cat:head:vform:bse)).
\end{verbatim}
The \texttt{cons} and \texttt{goal} operators are defined by ALE, and
used to separate the type antecedent from the description component of
the consequent (in this case, just the variable, \texttt{X}), and the
description component of the consequent from its relational
attachment.  We know that any feature structure subsumed by the
original antecedent will also be subsumed by the most general feature
structure of type \emph{sign}, because \emph{sign} introduces
\textsc{synsem}.

\subsection{Reducing Complex Conditionals}

The next step is to decompose \texttt{fswhen/2} itself.  For
simplicity, it can be assumed that its first argument is actually
drawn from a more general conditional language, including those of the
form $V=\mathit{Desc}$ and closure under conjunction and disjunction.
Also for simplicity, it is assumed that the variables of each
$\mathit{Desc}_i$ are distinct.

Such a complex conditional can easily be converted into a normal form
in which for each atomic conditional, $V_i=\mathit{Desc}_i$,
$\mathit{Desc}_i$ is non-disjunctive.  Just as with the principles
themselves, we can begin by isolating types and assuming that we have
a predicate \texttt{typewhen(Type,V,Goal)} that delays \texttt{Goal}
until \texttt{V} is of type \texttt{Type}.  We can then convert all of
the other conditionals to this by converting \texttt{fswhen(C,Goal)}
to \texttt{Conv}, i.e., \texttt{reduce(C,Goal,Conv)}, where
\texttt{reduce/3} is defined such that:
\begin{verbatim}
reduce((VD1,VD2),Goal,Conv) :-    % fswhen((VD1,VD2),Goal) iff
  reduce(VD2,Goal,Goal2),         %  fswhen(VD1,fswhen(VD2,Goal))
  reduce(VD1,Goal2,Conv).

reduce((VD1;VD2),Goal,(Conv1,Conv2)) :-
  reduce(VD1,(prolog(Trigger = 0) -> Goal ; true),Conv1),
  reduce(VD2,(prolog(Trigger = 1) -> Goal ; true),Conv2).

reduce(V=Desc,Goal,Conv) :-
  vars_of(Desc,Vars),
  desc_reduce(Desc,V,Goal,Vars,_,Conv).
\end{verbatim}
and \texttt{desc\_reduce/6} is defined such that
\texttt{desc\_reduce(Desc,V,Goal,Vars,VarsRest,Conv)} holds if and
only if \texttt{fswhen(V=Desc,Goal)} reduces to \texttt{Conv} and
\texttt{Vars-VarsRest} is the collection of all \texttt{Vars} that
occur in \texttt{Desc}:
\begin{verbatim}
desc_reduce(X,V,Goal,Vars,VarsRest,Conv) :-
  var(X),
  !,( select(Vars,X,VarsRest) -> Conv = (V = X,call(Goal))
      % if first occurrence of X, bind to V

    ; Conv = when(?=(V,X),((V==X) -> call(Goal) ; true)),
      VarsRest = Vars
      % otherwise, wait until V==X
    ).

desc_reduce(F:Desc,V,Goal,Vars,VarsRest,
            typewhen(IntroType,V,(farg(F,V,FVal),
                                  call(DescConv)))) :-
  intro(F,IntroType),
  desc_reduce(Desc,FVal,Goal,Vars,VarsRest,DescConv).

desc_reduce(T,V,Goal,Vars,Vars,Conv) :-
  type(T),
  !, Conv = typewhen(T,V,Goal).

desc_reduce((Desc1,Desc2),V,Goal,Vars,VarsRest,Conv2) :-
  desc_reduce(Desc1,V,Goal,Vars,VarsMid,Conv1),
  desc_reduce(Desc2,V,Conv1,VarsMid,VarsRest,Conv2).
\end{verbatim}
Here, the \texttt{prolog/1} goals are inserted without modification by
a compiler, as goals in that position are normally interpreted to be
relations over feature structure descriptions.  The Prolog convention
of comma for AND and semi-colon for OR is also used in these
definitions.  In the disjunctive case of \texttt{reduce/3}, the
binding of the variable \texttt{Trigger} is necessary to ensure that
Goal is only resolved once in case the goals for both conditionals
eventually unsuspend.  In the variable case, \texttt{desc\_reduce/6}
simply binds the feature structure in question when it first
encounters a variable, but subsequent occurrences of that variable
create a suspension that compiles directly down to Prolog
\texttt{when/2}, checking for identity with the previous occurrences.
Notice also that $\mathit{Intro}(\mathsc{f})$, here called
\texttt{IntroType}, can be used as the type to delay on in the case of
a feature value description. \texttt{farg(F,V,FVal)} binds
\texttt{FVal} to the argument position of \texttt{V} that corresponds
to the feature \texttt{F} in the term encoding of that feature
structure once \texttt{V} has been instantiated to a type for which
\texttt{F} is appropriate.

In the running example, we must convert the large \texttt{fswhen/2}
relational attachment to simpler \texttt{typewhen/2} goals.  The
second clause of the relation, \texttt{desc\_reduce/6} tells us that
this can be achieved by successively waiting for the types that
introduce each of the features, \textsc{synsem}, \textsc{loc},
\textsc{cat}.  Those types are \emph{sign},
\emph{syntax\_semantics} and \emph{local}, respectively:
\newpage
\begin{verbatim}
sign cons X 
     goal typewhen(sign,X,(farg(synsem,X,SynVal),
          typewhen(syntax_semantics,SynVal,(farg(loc,SynVal,LocVal),
          typewhen(local,LocVal,(farg(cat,LocVal,CatVal),
          fswhen((CatVal=(head:verb, marking:fin)),
                 (X=synsem:loc:cat:head:vform:bse)))))))).
\end{verbatim}
In practice, a great deal of static analysis is possible to reduce the
complexity of the resulting relational goals.  In this case, static
analysis of the type system tells us that since \emph{sign} is the
trigger type, \emph{syntax\_semantics} is the least appropriate type
of a \textsc{synsem} value, and \emph{local} is the least appropriate
type of a \textsc{loc} value, all three of these \texttt{typewhen/2}
calls can be eliminated:
\begin{verbatim}
sign cons X goal (farg(synsem,X,SynVal),
                  farg(loc,SynVal,LocVal),
                  farg(cat,LocVal,CatVal),
                  fswhen((CatVal=(head:verb, marking:fin)),
                         (X=synsem:loc:cat:head:vform:bse))).
\end{verbatim}
The description that \texttt{CatVal} is suspended on is a conjunctive
description, so the last clause of \texttt{desc\_reduce/6} tells us
that we should successively suspend on each conjunct.  The type that
introduces both \textsc{head} and \textsc{marking} is \emph{category}.
\begin{verbatim}
sign cons X 
     goal (farg(synsem,X,SynVal),
           farg(loc,SynVal,LocVal),
           farg(cat,LocVal,CatVal),
           typewhen(category,CatVal,(farg(head,CatVal,HdVal),
           typewhen(verb,HdVal,
             typewhen(category,CatVal,(farg(marking,CatVal,MkVal),
             typewhen(fin,MkVal,(X=synsem:loc:cat:head:vform:bse))
             ))
           )))
          ).
\end{verbatim}
Since \emph{category} is the least appropriate type of a \textsc{cat}
value, static analysis again allows us to simplify the code:
\begin{verbatim}
sign cons X 
     goal (farg(synsem,X,SynVal),
           farg(loc,SynVal,LocVal),
           farg(cat,LocVal,CatVal),
           farg(head,CatVal,HdVal),
           typewhen(verb,HdVal,
             (farg(marking,CatVal,MkVal),
              typewhen(fin,MkVal,(X=synsem:loc:cat:head:vform:bse))
             )
           )
          ).
\end{verbatim}

\subsection{Compiling Delays on Types}

Delaying until a feature structure becomes at least as specific as a
certain type is not so easy because of appropriateness conditions.
The term encoding of a feature structure may actually reflect the type
in the conditional before all of its values have been combined to
create a well-formed feature structure due to the check that must be
performed to enforce constraints.  CHR allows one very simply to
suspend until subsumption by an entire term is achieved.  We define a
CHR constraint \texttt{typewhen/3}, such that for each
\texttt{Type}, we add a rule:
\begin{verbatim}
typewhen(Type,MGSat,Goal) <=> call(Goal).
\end{verbatim}
where \texttt{MGSat} is the most general satisfier of \texttt{Type}.
This constraint, when posted by the calls in \texttt{desc\_reduce/6}
above, will wait until the variable in question is bound to something
at least as specific as \texttt{MGSat}, which is a term encoding for
which all appropriateness conditions are satisfied.
\texttt{Type} sits in the first argument position in order to index
the constraint into the correct rule more quickly.

In the running example, the term encoding of the most general
satisfier of \emph{sign} is:
\begin{quote}
\texttt{'\$sign'(\_,\_,'\$mod\_synsem'(syntax\_semantics,\_,\_,\_),\_)}.
\end{quote}
The first three argument positions correspond to the three features
that are appropriate to \emph{sign}, the first two of which have not
been mentioned here and, again by static analysis, do not need to be
instantiated.  The third, however, is the \textsc{synsem} value, and
the subterm in that position is the most general satisfier of the
type, \emph{syntax\_semantics}.  The CHR compiler will create term
subsumption code that causes the rule handler to wait for this feature
value to be instantiated to at least this subterm, and thus satisfy
the appropriateness conditions of \emph{sign}.  The last argument
position is always an anonymous variable and preserves the
intensionality of the logic.  The \texttt{typewhen/2} clause for
\emph{syntax\_semantics} is thus:
\begin{verbatim}
typewhen(sign,
         '$sign'(_,_,'$mod_synsem'(syntax_semantics,_,_,_),_),Goal)
 <=> call(Goal).
\end{verbatim}

This, of course, only uses a small part of CHR's functionality, but it
has proven to be a vital part for combining constraint resolution,
coroutining, appropriateness and logic programming in the logic of
typed feature structures.  It is actually possible to use
\texttt{when/2} directly to achieve the same effect, but in principle,
CHR should be more efficient, because it could identify at the
abstract machine level when subsumption fails, thus eliminating the
constraint from the store, as soon as possible.  Finding the right
order in which to check the various components of a term encoding
directly with \texttt{when/2} is not at all trivial.  This point is
discussed further in Section~\ref{desiderata:sec}.

\section{Subtype Covering Constraints}
\label{subcover:sec}

Another use of CHR as a very sophisticated suspension tool arises with
the enforcement of a different kind of internal constraint within
feature-structure-based grammar development systems.

The logic of typed feature structures is a logic with subtyping, and
this ordering among types induces an ordering among feature
structures, called \emph{subsumption}.  Our view of language, however,
is one of utterances that are maximally informative, and therefore
discretely ordered.  As a result, what we actually care about
determining in the context of grammar development is whether there is
a \emph{maximally informative} feature structure for a given utterance
with a non-empty denotation in our realist model of language.  Maximal
feature structures are those in which every node is assigned a
maximally specific type, and every pair of nodes with identical types
is either re-entrant or explicitly inequated.
\footnote{An inequation is a negative constraint that prohibits
re-entrancies, much as in Prolog II.  Their implementation is not
discussed in detail here, but they can essentially be reduced
to an underlying Prolog's \texttt{dif/2} statement using the term
encoding mentioned above.}
Non-maximally-specific types can simply be interpreted as a short-hand
for sets of their maximally specific subtypes, and non-maximal feature
structures can thus be viewed as a very compact kind of representation
for disjunctions of maximal feature structures.

Because of appropriateness and re-entrancies, not every feature
structure has a maximal extension.  Suppose type $a$ has two
appropriate features, \textsc{f} and \textsc{g}, both of whose values
are restricted to the type \textit{polarity}, which has maximal
subtypes $\mathord{+}$ and $\mathord{-}$.  $a$ could have exactly two
subtypes, $b$ and $c$, with appropriateness conditions such that $b$
must have an \textsc{f} value of $\mathord{+}$ and a \textsc{g} value
of $\mathord{-}$, and $c$ must have an \textsc{f} value of
$\mathord{-}$ and a \textsc{g} value of $\mathord{+}$.  In this case,
the non-maximal feature structure:
\begin{center}
\begin{avm}
\[ \rm a\\
f\; \@1 \rm bool\\
g\; \@1
\]
\end{avm}
\end{center}
has no maximal extension because $\mathord{+}$ and $\mathord{-}$ are
inconsistent.  Situations such as these do occasionally
arise in the realm of linguistic knowledge representation, and a
grammar development system based on feature structures must detect
them early in order to fail as soon as possible.

These can be implemented in ALE by adding relational attachments to
ALE's type-antecedent universal constraints that will suspend a goal
on candidate feature structures with types such as $a$ that could have
this problem (called \emph{deranged types}).  The suspended goal
unblocks whenever the deranged type or the type of one of its
appropriate features' values is updated to a more specific subtype,
and checks the types of the appropriate features' values.  CHR's
ability not only to suspend, but to suspend until a particular
variable is instantiated or even bound to another variable is the
powerful kind of mechanism required to check these constraints
efficiently, i.e., only when necessary.  Re-entrancies in a term
encoding may only show up as the binding of two uninstantiated
variables, and re-entrancies are an important case where these
constraints need to be checked.

We can create a constraint, \texttt{subtype\_cover/2}, whose first
argument is a deranged type.  For each such type, we can add three
kinds of rules.  The first retrieves the current type of a feature
structure from its term encoding and compares it to a deranged type
that it used to have.
\begin{verbatim}
subtype_cover(Type,FS) <=> 
   true & (type_index(FS,FSType),FSType \== Type)
 | true.
\end{verbatim}
In the case of our example with $a$, $b$ and $c$, $a$ is deranged, so
the following constraint handling rule is generated:
\begin{verbatim}
subtype_cover(a,FS) <=> 
   true & (type_index(FS,FSType),FSType \== a)
 | true.
\end{verbatim}
If a feature structure of type $a$ is refined to be a well-formed
feature structure of type $b$, then it is no longer in danger
(unification over the term encoding itself fails on appropriateness
failures, i.e., non-empty denotations, in such a case).  In fact, no
maximally specific type can be deranged.  If the type has changed,
then if it is still deranged, it is the responsibility of the rules
for the new deranged type to check it.

The second kind of rule compares a feature structure's appropriate
feature values to a product of types that belong to one of the
deranged type's ``safe'' subtypes (such as its maximally specific
subtypes).  If this product subsumes the types of the feature
structure's current values, then it is out of danger.  So for each
such product, there is one rule of the form:
\begin{verbatim}
subtype_cover(Type,AppropFeatProduct) <=> true.
\end{verbatim}
In the case of $a$, the following rules are generated:
\begin{verbatim}
subtype_cover(a,'$a'('$polarity'(+,_),'$polarity'(-,_),_))
 <=> true.
subtype_cover(a,'$a'('$polarity'(-,_),'$polarity'(+,_),_))
 <=> true.
\end{verbatim}
The first rule contains a term encoding of the feature structure:
\begin{center}
\begin{avm}
\[ \rm a\\
f\; \rm +\\
g\; \rm -
\]
\end{avm}
\end{center}
Its feature values' types correspond to the appropriateness conditions
for $b$.  The second rule's feature values' types correspond to the
appropriateness conditions for $c$.  In either case, if a feature
structure of type $a$ has these, then it is guaranteed to have a
maximally specific extension and thus denote in the language we are
modeling.

Finally, it could be that the feature structure already has no maximal
extensions, or only one maximal extension.  In the former case, we
should fail, and in the latter case, we should extend the feature
structure since that is the only possible extension it can have.  The
following rule uses a guard to count how many extensions are still
consistent with the current feature structure: 0, 1 or more than 1.
If there is more than 1, it fails, resuspending the constraint --- we
may still need to check later because it could lose consistency with
one of those extensions.  If it has 0 or 1, the guard succeeds and the
body of the rule fails or extends the feature structure through
unification as appropriate.
\begin{verbatim}
subtype_cover(Type,FS) 
               <=> true & stc_unify_test(Type,FS,N)
                   | ( N == 0 -> fail              
                     ; % N == 1
                       stc_unify_product(Type,FS)).
\end{verbatim}

\section{Further Desiderata}
\label{desiderata:sec}

There are two significant shortcomings to the view of logic
programming with typed feature structures presented here.  The first
is that feature structure unification, by far the most common
operation performed by a parser or generator based on this logic,
cannot always be reduced to Prolog unification. The reason, as
mentioned earlier, is that certain checks must be made in order to
detect whether to apply the consequent of a ``universal constraint,''
a constraint that applies to literally every typed feature structure,
although restrictions such as limiting the antecedent to a type and
applying constraint consequences under subsumption ameliorate the cost
of that task.  The view of constraints in both \texttt{when/2} and CHR
is that a constraint is posted relative to one or more terms and only
then is its satisfaction tracked by the underlying abstract machine.
What grammar developers (more specifically, people who write grammar
development systems) need right now is the ability to state that a
constraint universally applies, and apply that constraint to terms /
feature structures efficiently at the abstract machine level.  This is
the only level at which it will not be necessary to disrupt term
unification as ALE currently must.

The second shortcoming is the efficiency of the current CHR
implementation in SICStus Prolog.  This relies crucially on an
attributed variables package that is rather inefficient.  That
inefficiency is what in fact prevents grammar developers working with
feature structures from reducing those feature structures directly to
attributed variables, which are at least cosmetically quite similar.
This is in part due to the fact that there is no notion of
appropriateness nor a strong enough notion of typing to support
appropriateness, and thus encode typed feature structures as
efficiently as a term encoding currently can.  Simply looking at its
application within CHR, however, even there, the number of allowable
constraints in a handler does not scale up well, and even with a small
number of constraints, delaying works far too slowly.  As mentioned
earlier, one can implement \texttt{typewhen/2} directly in terms of
\texttt{when/2} at the cost of possibly having an overwhelming number
of residual suspensions due to an unfortunate ordering of atomic
suspensions that collectively implement delaying on subsumption by an
entire term.  Even with this large number of residues, our experiments
with a large HPSG grammar for German show that a direct implementation
in terms of \texttt{when/2} in SICStus Prolog is faster than using
SICStus's CHR library by a factor of 60 or more on our test suites.
In principle, CHR should be much faster.  Currently, the SICStus CHR
compiler simply partially evaluates the necessary run-time subsumption
checks, where possible, at the Prolog level.

\section{Conclusion}

The compilation of two kinds of constraints was proposed relative to a
preprocessor for a logic programming language that term-encodes typed
feature structures for a very transparent treatment of unification and
constraint logic programming with feature structures within Prolog.
CHR has essentially retired the problem of how best to handle the
delaying of constraints within grammar development systems, although
there remain a few bottlenecks to efficiency such as the universal
quantification implicit to linguistic constraints and in the SICStus
Prolog implementation, the underlying dependence of the CHR library on
the current implementation of attributed variables.

The extension of ALE described here, called TRALE, was developed at
the the University of T\"ubingen in parallel to a grammar of an
extensive fragment of German, both of which will be made publicly
available at the end of the year 2000 as deliverables of the
Sonderforschungsbereich 340.

\bibliography{thesis}

\end{document}